\documentclass{article}

\usepackage{microtype}
\usepackage{graphicx}
\usepackage{subcaption}
\usepackage{booktabs} % for professional tables
\usepackage{amsmath}
\usepackage{multirow}
\usepackage{listings}
\usepackage{xcolor}
\usepackage{tcolorbox}

\definecolor{codegray}{gray}{0.95}
\definecolor{commentgreen}{rgb}{0.0,0.45,0.0}
\definecolor{keywordblue}{rgb}{0.0,0.2,0.65}
\definecolor{stringred}{rgb}{0.55,0.0,0.0}

\lstdefinestyle{pythonstyle}{
    language=Python,
    backgroundcolor=\color{codegray},
    basicstyle=\ttfamily\footnotesize,
    keywordstyle=\color{keywordblue}\bfseries,
    commentstyle=\color{commentgreen}\itshape,
    stringstyle=\color{stringred},
    numbers=left,
    numberstyle=\tiny\color{gray},
    stepnumber=1,
    numbersep=6pt,
    frame=single,
    rulecolor=\color{gray},
    breaklines=true,
    breakatwhitespace=true,
    showstringspaces=false,
    tabsize=4,
    captionpos=b,
    keepspaces=true,
    columns=fullflexible
}

\newcommand{\ourtool}{\textbf{T2J\space}}
\newcommand{\ourtoolnospace}{\textbf{T2J}}

\usepackage{hyperref}

\usepackage[accepted]{icml2026}

\usepackage{amsmath}
\usepackage{amssymb}
\usepackage{mathtools}
\usepackage{amsthm}

\usepackage[capitalize,noabbrev]{cleveref}

\theoremstyle{plain}

\theoremstyle{definition}

\theoremstyle{remark}

\usepackage[textsize=tiny]{todonotes}

\icmltitlerunning{Learning Bug Context for PyTorch-to-JAX Translation with LLMs}

\begin{document}

\twocolumn[
  \icmltitle{Learning Bug Context for PyTorch-to-JAX Translation with LLMs}



  \icmlsetsymbol{equal}{*}

  \begin{icmlauthorlist}
    \icmlauthor{Hung Phan}{xxx}
    \icmlauthor{Son Vu}{xxx2}
    \icmlauthor{Tuan Dinh}{yyy}
    \icmlauthor{Nesreen Ahmed}{zzz}
    \icmlauthor{Ali Payani}{yyy}
    \icmlauthor{Ali Jannesari}{xxx}
   
  \end{icmlauthorlist}

  \icmlaffiliation{xxx}{Department of Computer Science, Iowa State University, Ames, IA, USA}
  \icmlaffiliation{xxx2}{Department of Human Science, Iowa State University, Ames, IA, USA}
  \icmlaffiliation{yyy}{Subquadratic AI, Miami, Florida, USA}
  \icmlaffiliation{zzz}{Cisco AI Research, Outshift, San Jose, CA, USA}

  \icmlkeywords{Machine Learning, ICML}

  \vskip 0.3in
]



\printAffiliationsAndNotice{}  % no special notice (required even if empty)

\begin{abstract}
% Code Translation
Large language models (LLMs) have shown strong performance on code translation between widely used programming languages. However, translation becomes much less reliable for domain-specific code, where correctness depends on framework-specific APIs and execution semantics. One example is translating deep-learning code from PyTorch to JAX, where LLM outputs often contain subtle bugs or non-idiomatic usage that prevents execution or changes behavior. Prior work suggests that curated bug–fix data from LLM-generated code can help improve code generation quality, but such resources are still limited for PyTorch-to-JAX translation.

In this work, we introduce \ourtoolnospace, a benchmark of LLM translation bugs paired with developer-written fixes for PyTorch-to-JAX code. We start from 20 kernels in the TorchLeet dataset, translate them to JAX using the weak LLM gpt-4o-mini, and hire software developers to debug and repair the generated JAX implementations. In total, \ourtool contains over 160 bug–solution pairs collected from real LLM outputs. We then use \ourtool to improve PyTorch-to-JAX translation for the weak LLM gpt-4o-mini via in-context learning. Our evaluation shows that using \ourtool yields up to 20\% improvement of our proposed metric $\ourtoolnospace\_CodeTrans\_Score$. We release our replication package at here\footnote{https://tinyurl.com/4ehky8hb}
% , demonstrating the value of developer-curated bug–fix supervision for more reliable framework-level code translation.

% Large Language Models have many success in general code translation between well-known programming languages. However, in the domain-specific languages code translation, there are risks of getting low quality translation results due to the lack of models' knowledge about the source/ target languages. Previous works show that collecting a dataset of bugs pattern from Large Language Models (LLM) generated code are needed to improve the quality of code generation on general programming languages. In this work, we select the research problem as code translation from PyTorch to JAX code. To overcome the challenge of lacking of bug-solution of LLM generated code for PyTorch-to-JAX translation, we collect a benchmark of LLM bugs and hired software developers to conduct fixing solution for each bugs appeared in LLM generated JAX code over 20 kernels of the well-known TorchLeet dataset in PyTorch. In total, we collect over 400 bugs and bugs' solutions from JAX generated code of two weak LLMs: gpt-4o-mini and QwenCoder-2.5-7B-IT. We use this benchmark, T2J-bench, to improve the quality in code translation of numerous open LLMs by two directions: in-context learning and fine-tuning with T2J-bench. Evaluation shows that T2J-bench can achieve up to XX\% of improvement in terms of ICE-score, which shows the potential of this benchmark in improving PyTorch-to-JAX code translation.

\end{abstract}

% \begin{IEEEkeywords}
% code generation, in context learning,cross-library code translation.
% \end{IEEEkeywords}

\section{Introduction}
% Application of LLMs in code translation (write 7-10 sentences).
% Challenges of cross-frameworks code translation by LLMs, such as Pytorch-to-JAX translation: Accuracy might not high.
% Challenges of code translation evaluation metrics for Pytorch-to-JAX translation.
% Current researches show that there are a lot of bugs in code translation with LLMs (motivate our research direction: creating the fixed-bug dataset.
% Overview of our approach, T2J. What modules do we have, and what are our contributions?
Code translation involves converting a program from one programming language to another while preserving the original functionality. This process is useful for cross-language and cross-domain migration, allowing organizations to transition their code base to more modern languages or to various purposes. It also supports the modernization of legacy systems by re-implementing them in languages that promote greater maintainability and scalability as part of system refactoring efforts . Furthermore, in enterprises that employ multiple programming languages, code translation enhances the productivity of the programmer. 

However, recent research work ~\cite{pan2024lostInTranslation}, ~\cite{dou2024whatswrongcodegenerated} indicates that LLMs-generated programs in the target language continue to encounter various quality problems, including compilation errors or functional inconsistencies. In response to these issues, Yang et al ~\cite{zhang2024uniTrans} developed UniTrans, which augments LLM-based code translation through an iterative correction mechanism. In particular, UniTrans utilizes LLMs to refine the translated program by incorporating test inputs and outputs or compilation error feedback. While demonstrating potential, UniTrans nonetheless fails to rectify the translated program in a considerable number of instances, particularly when relying on smaller LLMs (those with fewer than 10 billion parameters). For example, when integrated with LLaMA-7B, UniTrans only elevates translation accuracy from 31.25\% to 31.82\% (representing a mere 0.57\% improvement) for translations from Java to Python; and for translations from Python to C++ it boasts a translation accuracy improvement from 92.72\% to 94.43\% (barely 1.70\% leap).

\noindent These limitations become more profound in specific situations, such as translating PyTorch code/programs to JAX.  This is because of core differences in dynamic graph execution, just-in-time (JIT) compilation, automatic differentiation approaches, and vectorization flows. During PyTorch-to-JAX migrations, LLMs often struggle to handle tensor operations, gradient calculations, or device-free optimizations. This results in non-functional or even incorrect code that strays from the original intent of code. 

% \noindent\textbf{Our contribution.} To address these limitations, we introduce \textbf{T2J}, a novel framework for PyTorch-to-JAX code translation that integrates modular components for improved accuracy and dependability. T2J comprises three core modules: (1) a bug rectification module that utilizes our curated fixed-bug dataset to iteratively correct hallucinations and logical flaws through learning from human feedback. Our primary contributions include the creation of the first fixed-bug dataset for PyTorch-to-JAX translations, encompassing over 1,000 annotated pairs with bug classifications and resolutions; empirical demonstrations of T2J achieving up to 25\% higher translation accuracy compared to baseline LLMs; and a comprehensive benchmark suite for evaluating cross-framework migrations, fostering reproducible advancements in this domain

\noindent\textbf{Our contribution.} To address these challenges, we introduce T2J, a prompt augmentation framework designed to enhance LLM-based PyTorch-to-JAX code translation by leveraging curated datasets and structured prompting strategies. 
The framework proceeds in several key stages: first, we construct parallel corpora from established PyTorch datasets, in particular TorchLeet and CodeParrot. Then we employ high-quality GPT models such as GPT-4o to produce initial JAX translations. Subsequently, professional human developers iteratively refine the translated JAX program to achieve functional equivalence with the original PyTorch input, producing a curated fixed-bug dataset that systematically documents prevalent translation errors—such as API mismatches, gradient computation discrepancies, and optimization failures—and their corresponding resolutions. The dataset also includes error-by-error fix instruction, and also. Building on this, we design augmented prompts that integrate targeted, structured guidance derived from the fixed-bug dataset. Finally, we evaluate T2J performance across both datasets using the CodeBLEU metric, alongside three novel metrics—Fixing Cost (quantifying the effort required for post-translation corrections), CodeTran\_Score (leveraging LLM-as-a-judges for evaluating code usefulness and functional correctness), and Comparison-Score (leveraging LLM-as-a-judges for binary judgement)—to provide a comprehensive assessment of translation quality. We contribute (1) the creation of the first fixed-bug dataset specifically for PyTorch-to-JAX translations, encompassing detailed annotations of error patterns and fixes to facilitate improvements in reliability on LLM code translation; (2) the T2J framework, which innovates prompt augmentation techniques to bridge domain-specific gaps, carving path for a scalable approach for other cross-ecosystem migrations; (3) rigorous empirical validation showing substantial performance enhancements of our framework.

The remainder of this paper is structured as follows. Section 2, Motivation Example, presents a case study demonstrating the use of LLMs for PyTorch-to-JAX translation. Section 3, Background, introduces key concepts relevant to our work. Section 4 describes the core modules of the T2J framework. Section 5 discusses the central research questions and presents evaluation results. Finally, we summarise related work, limitations and provide future directions in the last three sections.

% \noindent Significant challenges still exist and persist among 
% cross-frameworks and cross-domain translation ~\cite{tao2024unravelingpotentiallargelanguage}. These issues, which cause problems such as suboptimal accuracy, often result from complexities in handling framework-specific and domain-specific abstractions, semantic discrepancies, and performance optimizations. LLMs frequently encounter difficulties with syntactic variations, API mismatches, incorrect data types, or logical errors, which make translated code incorrect, particularly in scenarios involving code translations across context and programming languages. 
% Additionally, these models are prone to hallucinations, generating non-syntactic "code bugs" that execute without immediate errors but produce incorrect outputs, as demonstrated in empirical studies across various translation tasks. 
% Evaluating such translations, being a daunting labor task itself, also presents technical hurdles, as traditional metrics may fail to adequately capture runtime behaviors, edge cases, and contextual dependencies, often leading to an underestimation of errors in complex codebases. Recent research highlights the prevalence of bugs in LLM-generated code, including syntax errors, incomplete implementations, and vulnerabilities, with analyses identifying numerous distinct bug patterns across prominent models.
\begin{figure*}[h]
  \centering
  \begin{minipage}[t]{0.30\textwidth}
    \centering
    \textbf{(a) Input PyTorch Code}
    \begin{lstlisting}[style=pythonstyle]
import torch
import torch.nn as nn
import torch.nn.functional as F

# Define model
class SimpleNN(nn.Module):
    def __init__(self):
        super(SimpleNN, self).__init__()
        self.fc = nn.Linear(28 * 28, 10)

    def forward(self, x):
        x = x.view(-1, 28 * 28)  # Flatten
        return self.fc(x)

# Example
model = SimpleNN()
input_tensor = torch.randn(1, 1, 28, 28)
output = model(input_tensor)
print(output)
    \end{lstlisting}
  \end{minipage}
  \hfill
  \begin{minipage}[t]{0.30\textwidth}
    \centering
    \textbf{(b) Incorrect JAX code}
    \begin{lstlisting}[style=pythonstyle]
import jax
import jax.numpy as jnp
from jax import random

# Initialize parameters 
def init_params(key):
    w_key, b_key = random.split(key)
    # Wrong shape
    W = random.normal(w_key, @(10, 28*28)@) * jnp.sqrt(1 / (28*28))
    b = jnp.zeros((10,))
    return {'W': W, 'b': b}

# Forward function
def forward(params, x):
    x = x.reshape(x.shape[0], -1)
    return jnp.dot(x, params['W']) + params['b']  

# Example
key = random.PRNGKey(0)
params = init_params(key)
input_tensor = random.normal(key, (1, 1, 28, 28))
output = forward(params, input_tensor)
print(output)
    \end{lstlisting}
  \end{minipage}
  \hfill
  \begin{minipage}[t]{0.30\textwidth}
    \centering
    \textbf{(c) Correct JAX Code}
    \begin{lstlisting}[style=pythonstyle]
import jax
import jax.numpy as jnp
from jax import random

# Initialize parameters
def init_params(key):
    w_key, b_key = random.split(key)
    # Correct shape
    W = random.normal(w_key, ^(28*28, 10)^) * jnp.sqrt(1 / (28*28))
    b = jnp.zeros((10,))
    return {'W': W, 'b': b}

# Forward function
def forward(params, x):
    x = x.reshape(x.shape[0], -1)
    return jnp.dot(x, params['W']) + params['b']

# Example
key = random.PRNGKey(0)
params = init_params(key)
input_tensor = random.normal(key, (1, 1, 28, 28))
output = forward(params, input_tensor)
print(output)
    \end{lstlisting}
  \end{minipage}
  \caption{Example of PyTorch-to-JAX translation. (a) Input code; (b) Incorrect translation by 4o-mini; (c) Correct code.}
  \label{t2j:fig:motivation_example}
\end{figure*}

\section{Motivation Example}
An illustration of how low-cost LLM as 4o-mini generate JAX buggy code from Pytorch code can be shown in Figure \ref{t2j:fig:motivation_example}. The input Pytorch program, which is a simple neural network, defines a single-layer feedforward network that flattens a 28×28 input image and applies a linear transformation to produce a 10-dimensional output tensor. The JAX code generated by 4o-mini has an extra function called $init\_params$, while in Pytorch the corresponding function was performed in the construction of the Neural Network object. We can see that the generated code consumed an error in Line 8 (see Figure \ref{t2j:fig:motivation_example}b)), where it incorrectly passed the shape of a linear object defined by the parameter $W$. The correct version of the generated code, shown in Figure \ref{t2j:fig:motivation_example}, requires one modification step to correct the argument passed onto the $random.normal$ JAX API call. This example shows specific challenge of Pytorch-to-JAX translation that even with this simple code snippet, low-cost LLM still failed to extract the correct code.

% The output code in JAX will then be fixed based on the evaluation of the developers and the debuggers to make sure correctness.

% This example shows the  limitations of direct LLM-based translation for the Torch-to-JAX pipeline.

\section{Background}
% \textbf{Applications of close LLMs for code translation.} 
\textbf{Direction of Prompt Augmentation.}Compared to open-source LLMs, closed-source LLMs developed by large organizations generally demonstrate higher accuracy in question/answering tasks across multiple domains~\cite{hanke2024openllmsnecessarycurrent, du2024llmsassistnlpresearchers}. 
In the context of code translation, existing studies have shown that closed LLMs outperform open LLMs across a variety of programming languages~\cite{yan2023codetransoceancomprehensivemultilingualbenchmark}. To the best of our knowledge, the most advanced and widely adopted closed LLMs are the OpenAI's GPT models ~\cite{openai2025}. A major challenge in optimizing GPT models for code translation lies in the high computational cost of traditional machine learning techniques such as pre-training and fine-tuning. As a result, prompt augmentation has emerged as a practical and cost-effective alternative to improve closed LLM performance. One recent work in this direction is UniTrans~\cite{zhang2024uniTrans}, which introduces a framework for prompt augmentation using LLM-generated test cases, specifically targeting translations among widely used languages such as Python, C++, and Java. In our work, we focus on translating Python code between machine learning frameworks, specifically from PyTorch to JAX. We hypothesize that existing LLM-generated test cases may be less reliable in this setting due to the niche nature of such translations. To address this, we propose a mechanism for augmenting prompts with common error patterns and their corresponding workaround solutions, tailored to the PyTorch to JAX translation task. Supporting our motivation, Pan et al.~\cite{pan2024lostInTranslation} presents evidence that translated code across major programming languages often contains diverse types of errors. In this paper, we call the direction that we augment the prompt for Pytorch-to-JAX code translation as Fixed-bugs Prompt Augmentation.

\textbf{Optimizing Metrics for Code Translation Evaluation.} Traditional metrics for evaluating translated code often rely on the availability of a reference or "oracle" code. In our study, we use CodeBLEU \cite{ren2020codebleumethodautomaticevaluation} as a representative traditional metric to assess the quality of PyTorch-to-JAX translations. With the rise of Large Language Models (LLMs), however, these models are increasingly utilized not only for generating code translations but also for evaluating them. To the best of our knowledge, LLM-based evaluation offers two key advantages. First, such methods can function with or without reference code. In these cases, LLMs can generate synthetic labels to serve as pseudo ground truths for comparing translated outputs. Second, LLMs are capable of assessing multiple dimensions of code quality, including functional correctness and readability \cite{weyssow2024codeultrafeedbackllmasajudgedatasetaligning}. One of the most recent LLM-based evaluation techniques is ICE-score \cite{zhuo-2024-ice}, proposed by Zhuo et al., which targets natural language-to-code generation. In this framework, the authors define evaluation criteria and a scoring rubric to evaluate Python code based on two dimensions: functional correctness and usefulness. They employ GPT models, guided by specially designed prompts and rubrics, to assign scores ranging from 1 to 5. Building on ICE-score, we propose two optimization strategies tailored to code-to-code translation. First, we design customized prompts and scoring rubrics specifically for PyTorch-to-JAX translation, which we term \textbf{\ourtoolnospace\_CodeTrans\_Score}. Second, we develop a separate set of prompts—agnostic to specific coding preferences—to enable direct comparison between two translated code snippets. The LLMs are asked to judge which translation is superior. We refer to these as \textbf{Code Comparison Prompts}.

\section{Approach}
% https://docs.google.com/drawings/d/1UQNdF-QWB1HIGRBeSR2pFbd2VyG1WC8ZFHmAkl2attM/edit
\begin{figure*}[t]
    \centering
    \includegraphics[width=\linewidth]{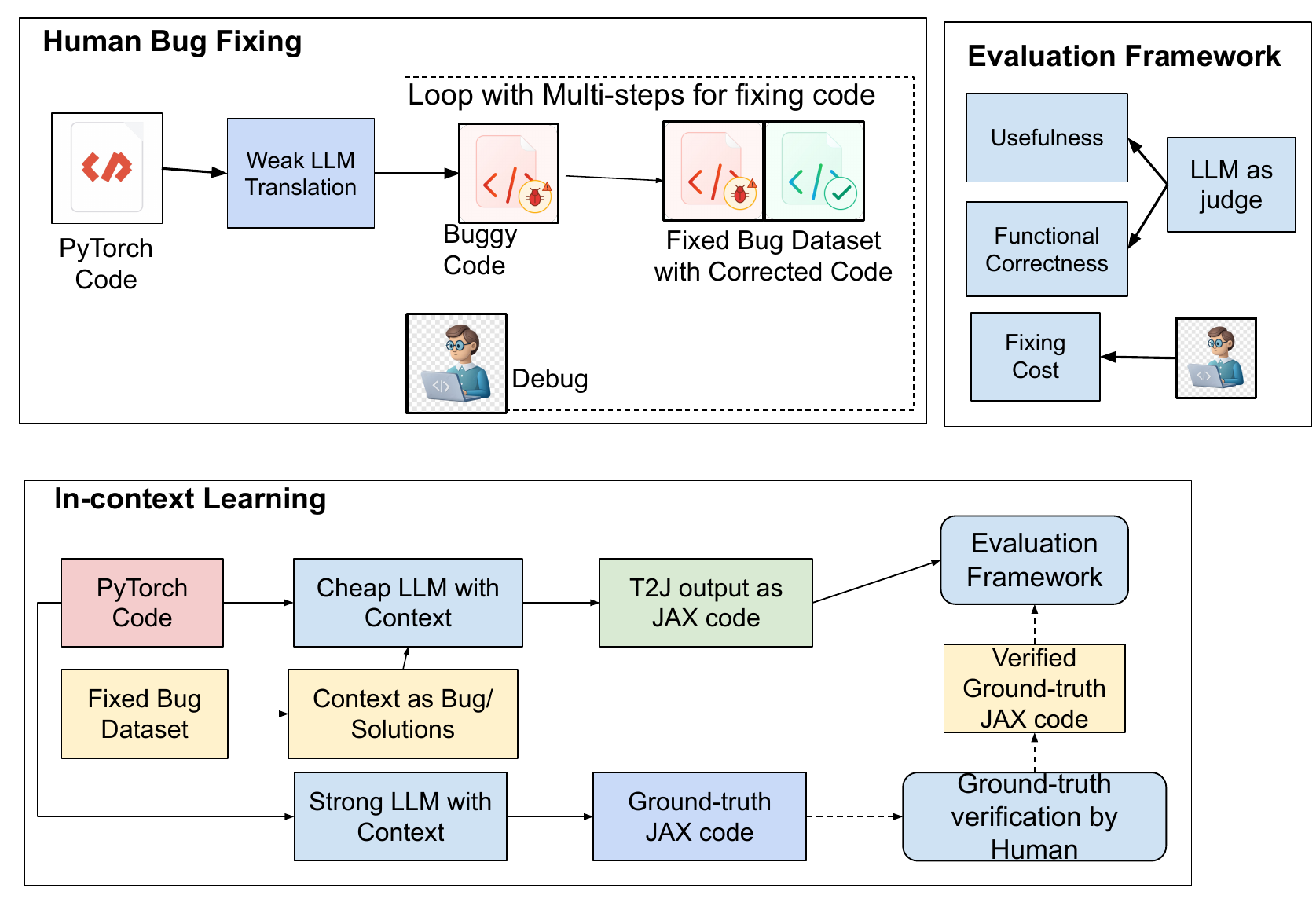}
    \caption{Overview Architecture of \ourtool. In Module One, a multi-step bug fixing process for the problem-solving dataset's code snippet was performed. In Module Two, Evaluation Framework, we proposed metrics for this domain code translation evaluation.  In module Three, we measure the accuracy of translation with different settings using an intrinsic dataset of 20 problem-solving code snippets and an extrinsic dataset of 100 randomly selected PyTorch code snippets selected from Github corpus. }
    \label{t2j:fig:architecture}
\end{figure*}
We depict modules of \ourtool in Figure \ref{t2j:fig:architecture}. In the first module, we describe how we hire software developers to check and modify JAX-generated code to ensure it is correct. We also define the definition of correct generated code in our work. The second module introduces our proposed metrics for comparing between predicted and expected generated code, in terms of using LLMs and leveraging the human bug fixing process's output. In the third module, we go in to details about the fixed bug dataset, which we will use for in-context learning with data about fixed bugs as extra context. In the fourth and fifth sections, we describe how \ourtool performs translation and how we generate the ground-truth dataset for evaluation. We also describe what baseline setting we use for comparison with our proposed pipeline. We call these modules intrinsic evaluation and extrinsic evaluation, and these modules are performed on two different datasets. Next, we discuss about important concepts and design selection we use for \ourtool.
\subsection{Design Selection}
\subsubsection{Selection of PyTorch Dataset}
We curate datasets of PyTorch code for two tasks. First, we create a dataset of fixed bugs as JAX-generated code from PyTorch input code for our proposed in-context learning process. This process requires the involvement of software developers and LLM for PyTorch-to-JAX code translation. The second task is the evaluation process, where the PyTorch code will be used as input for baseline models or our proposed translation framework to get the output as JAX code snippets for further processing to correct the code and evaluation. Depending on the tasks, we collect the PyTorch dataset from two domains: problem-solving code and general-purpose code.

\textbf{Problem-Solving Code Dataset.} We construct a PyTorch dataset based on popular coding interview problems implemented in PyTorch. Owing to the popularity of these problems, we assume that developers can readily verify solutions and fix bugs by consulting existing online resources. For this purpose, we leverage the TorchLeet dataset \citep{torchleet2025} as our problem-solving code corpus. We chose this dataset because each problem is scoped at the file level and because TorchLeet has been highly ranked by GitHub users. From this dataset, we collect all \textbf{20} code snippets, covering three difficulty levels: easy, medium, and hard. All snippets are compilable and runnable using the default test cases included within the dataset.

% \textbf{Problem Solving Code Dataset.} We attempt to collect a PyTorch dataset for popular coding interview problems in PyTorch. Given the popularity of each coding problem, we assume that software developers can easily check and fix bugs by searching online resources. We leverage the TorchLeet dataset \citep{torchleet2025} to serve as problem-solving code dataset. The main reasons for choosing this dataset are that the scope of each coding problem is at the file level, and the TorchLeet dataset was highly ranked by GitHub users. From this dataset, we collect all \textbf{20} code snippets with 3 difficulty levels: easy, medium, and hard. All of these code snippets are compilable and runnable with default test case provided in the content of them.

\textbf{GitHub Code Dataset.} To further evaluate our proposed PyTorch-to-JAX translation approach on a broader range of PyTorch code, we consider code snippets drawn from high-quality repositories on GitHub. Specifically, we use the PyTorch subset of the large-scale CodeParrot dataset \cite{codeparrot2022} as our second evaluation corpus. From this dataset, we extract \textbf{100} PyTorch code snippets. Unlike the Problem-Solving Code dataset, the GitHub PyTorch snippets originate from diverse repositories and are not guaranteed to be directly compilable.

% \textbf{Github Code Dataset.} We attempt to evaluate our proposed approach in PyTorch-to-JAX translation on larger scope of PyTorch code snippet that can belong to high quality code repositories on Github. We use the PyTorch subset of code snippets provided in the large scale code dataset CodeParrot \cite{codeparrot2022} as the second dataset for our evaluation. We extract \textbf{100} code snippets in PyTorch for evaluation. Different from Problem Solving Code dataset, the Github Pytorch code belongs to multiple code repositories and cannot be guaranteed to be compilable.
\subsubsection{Selection of LLMs}
\textbf{Cheap (Weak) LLMs.} We define low-cost LLMs as the target models to improve in the PyTorch-to-JAX translation task. These models, referred to as \textit{Cheap LLMs}, can be used without commercial API keys or additional costs. Moreover, we leverage their JAX-generated code to construct the fixed-bug dataset, under the assumption that cheap LLMs produce a higher proportion of buggy code, which can enrich this dataset. For our experiments, we select GPT-4o-mini, the least expensive model offered by OpenAI, as the representative cheap LLM.
% \textbf{Cheap LLM} We consider low-cost LLM as the target LLM to improve in PyTorch-to-JAX translation. These LLMs, called cheap LLM, can be run without any cost/ cormercial API keys. Besides, we also use the JAX generated code from cheap LLM to create the fixed bug dataset with the assumption that cheap LLM will generate more buggy code that can contribute to the fixed bug dataset. We select the cheap LLM as GPT-4o-mini, the cheapest model provided by OpenAI.

\textbf{Costly (Strong) LLMs.} We define high-cost LLMs, referred to as \textit{Costly LLMs}, as the source of ground-truth JAX code for comparison with the translations produced by cheap LLMs. We employ the GPT-4o model as the costly LLM. GPT-4o is one of the most widely used models provided by OpenAI.

% \textbf{Costly LLM.} We leverage high-cost LLM, called costly LLM, to get the ground-truth JAX code for comparison with the JAX code generated by weak LLM. In our setting, we use GPT-4o model as the costly LLM. GPT-4o has been considered as one of the most popular used model from OpenAI.

\subsubsection{Prompts}
We define these following types of prompt in our work.

\textbf{Standard Prompt.} We define the standard prompt as the basic prompt for translating code snippet from Pytorch to JAX (see Appendix A.1). In this prompt, we define the translation request by annotating the role of LLM as an expert in code translation and provide a basic request that translate from source language as PyTorch to target language as JAX code. 

\textbf{Augmented Prompt.} We design augmented prompt with following information. First, similar to the standard prompt, the augmented prompt will have information about the role of LLM and the requirement about input and output. Differ from standard prompt, the augmented prompt will specify a hint for LLMs as the list of errors and errors' solution provided by the fixed bug dataset, uploaded in the JSON format as context of prompt. Definitions of each fields in the fixed bug dataset are also included in the prompt content (see Appendix A.2).

\textbf{Evaluation Prompt.} We propose two metrics that leverage LLMs for code-to-code translation. From our knowledge, there are yet existing LLM prompt for doing this task. We define a set of prompts, called evaluation prompts (see Appendix A.3), to ask costly LLMs to evaluate source code in different aspects. We will describe in details of these metrics in the next section.

\subsubsection{Types of Evaluation}
\textbf{Intrinsic and Extrinsic Evaluation.} Our evaluation is divided into two parts. The first, referred to as intrinsic evaluation, is conducted on 20 problem-solving PyTorch code snippets using a cross-validation setting. The second, extrinsic evaluation, is performed on 100 samples of PyTorch code collected from GitHub. There are two key differences between these configurations. First, for intrinsic evaluation, the ground-truth JAX code is obtained through human verification and bug-fixing, whereas for extrinsic evaluation, we rely on JAX code generated by a costly LLM for the GitHub dataset.
% \textbf{Intrinsic Evaluation and Extrinsic Evaluation.} We set up two parts of the evaluation process. The first process, called intrinsic evaluation, provides an experiment on 20 code snippets of problem-solving PyTorch code snippets with cross cross-validation setting. The second process will perform the experiment on 100 samples of the GitHub PyTorch code. There are two different between configurations of intrinsic and extrinsic evaluation. First, while the ground truth JAX code generation was veriffied by human bug fixing process for intrinsic evaluation, we rely on costly LLM's JAX generated code over Github code dataset for performing extrinsic evaluation.

\subsection{Human Bug Fixing Process}
The core module of \ourtool is a systematic bug-fixing process designed to improve the reliability of LLM-generated translations from PyTorch to JAX. Starting with Python code written in PyTorch and the corresponding JAX-generated code from LLMs, this process produces a parallel dataset that pairs the original PyTorch snippets with their corresponding corrected JAX translations. To ensure correctness, we employed two professional software developers with over five years of Python programming experience to analyze and fix the outputs produced by LLMs. The JAX translations are then subjected to careful manual verification: the developers perform multiple rounds of debugging and correction until the translated JAX code passes all test cases and produces results equivalent to the original PyTorch implementation. During this stage, the verifiers execute the translated and fixed code using Python compilers to confirm correctness. At the end of the process, two complementary datasets are created: one containing pairs of PyTorch snippets and their validated JAX counterparts, and another capturing the bugs identified in LLM outputs along with their corresponding fixes. Importantly, this methodology is flexible, as we apply the same procedure to different PyTorch datasets and experiment with different LLMs depending on the objectives of other modules within our framework.

\textbf{Verified Code Correctness by Human.} We consider a fixed version as JAX code snippet verified by a human as correct if and only if the JAX-generated code can be compiled, runnable, and returns the same output as its corresponding Pytorch code snippet, given the same test case. In our work, the human bug fixing process was performed in the problem-solving dataset only since its code snippets have test case and can be runnable which we can rely on their execution output for comparison with the JAX generated code. 

\textbf{Fixed Bug Dataset.}
Given the input of 20 PyTorch samples from the TorchLeet dataset, two professional developers are hired to modify the code snippets generated from the LLM code translation process. The output of this process for each generated code is a set of multiple fix steps. We store this data in JSON array format to be usable as the context for \ourtool prompting technique. The fixed bug dataset was constructed by our selected cheap LLM 4o-mini. In total, this dataset contains \textbf{163} pairs of bugs/ solutions to fix bug.

\subsection{Evaluation Framework}
We use CodeBLEU \cite{ren2020codebleumethodautomaticevaluation}, a well-known code evaluation metric, as the baseline metric for PyTorch-to-JAX translation. Since CodeBLEU relies on the AST similarity between two code snippets, we assume that it cannot be a good metric for PyTorch-to-JAX translation output comparison, since both the JAX-generated code might be very similar to the ground truth code, such as in Motivation Example \ref{t2j:fig:motivation_example}. Recently, an LLM-based code evaluation metric has been proposed \citep{zhuo-2024-ice}. In this work, the authors proposed the ICE-score, a metric for natural language to code translation based on usefulness and functional correctness. In our work, we inherit the idea of the ICE-score and leverage our fixing process to propose three metrics for evaluating JAX-generated code. Given the $n$ pairs of Pytorch and JAX codes in the corresponding Pytorch code set P, the predicted code set JAX $J^{p}$, the ground truth (reference) code set JAX $J^{r}$ as ($p_{i}$,$j_{i}$), our proposing metrics will be provided as following.

\begin{table}[t]
\tiny
\centering
\caption{Evaluation preferences and their descriptions for \ourtoolnospace\_CodeTrans\_Score.}
\label{t2j:tbl:preference}
\resizebox{\columnwidth}{!}{%
\begin{tabular}{p{0.25\linewidth} p{0.7\linewidth}}
\toprule
\textbf{Preference} & \textbf{Description} \\
\midrule
Usefulness & How useful the JAX code is for replicating or adapting the functionality of a typical PyTorch source code implementation. \\
\addlinespace
Functional Correctness & How well the JAX code preserves the behavior of the original PyTorch code. You are to assess whether the JAX code would produce equivalent outputs to the original PyTorch code across possible inputs, even though the PyTorch code is not shown. Consider unit-test-style logic and general expectations of equivalence. \\
\bottomrule
\end{tabular}}
\end{table}

\textbf{\ourtoolnospace\_CodeTrans\_Score.} Similar to the ICE-Score metric \citep{zhuo-2024-ice} for natural language to code translation, we use the GPT-4o model to evaluate the quality of translated code by usefulness and functional correctness. We design prompt, called CodeTrans prompt, with corresponding evaluation criteria and scoring rubric from 0 (lowest) to 4 (highest). CodeTrans also be useable without having the reference code. Thus, we have a set of following metrics: \ourtoolnospace\_CodeTrans\_Use\_Ref (i.e. the metric for usefulness with reference), \ourtoolnospace\_CodeTrans\_Func\_Ref, \ourtoolnospace\_CodeTrans\_Use\_NoRef, \ourtoolnospace\_CodeTrans\_Func\_NoRef. Explanation of two aspects/ preferences is shown in Table \ref{t2j:tbl:preference}.

\textbf{\ourtoolnospace\_FixCost\_Score.} We measure the number of fix steps required to have the JAX correct code from input JAX initial translated code from LLM with this equation:

\begin{equation}
\begin{split}
\ourtoolnospace\_FixCost\_Score(J^{before\_fix},J^{correct})= \\
\frac{1}{n}\sum_{i=1}^n count\_fix\_step(j_{i}^{before\_fix},j_{i}^{correct})    
\label{t2j:eq:fixcost_score}
\end{split}
\end{equation}

In equation \ref{t2j:eq:fixcost_score}, $J^{before\_fix}$ is the set of JAX generated code as input for human verification process, while $J^{correct}$ is the final version of JAX fixed code that is correct, i.e. it can run and returns consistent output with its corresponding PyTorch code given the same input test case.

\textbf{\ourtoolnospace\_Comparison\_Score.} In this metric, we propose a direct comparison between two translation sets $J^{1}$ and $J^{2}$ to see which one is closer to the input PyTorch code set $P$, implemented by this equation:

\begin{equation}
\begin{split}
\ourtoolnospace\_Comparison\_Score(J^{1}, P, J^{2}) = \\
\frac{1}{n} \sum_{i=1}^n
\begin{cases}
1, & \text{if } \texttt{is\_better}(j_{i}^{1}, j_{i}^{2}, p_{i}) \\
0, & \text{otherwise}
\end{cases}
\end{split}
\label{t2j:eq:comparison_score}
\end{equation}

In equation \ref{t2j:eq:comparison_score},  the $\texttt{is\_better}(j_{i}^{1}, j_{i}^{2}, p_{i})$ function returns 1 if $j_{i}^{1}$ is considered as better code than $j_{i}^{2}$, given the input prompt for comparison called Comparison prompt (see Appendix A.3). Note, we also include the content of PyTorch code $p_{i}$ as a required input to help LLM comparing the input with each code candidate. The scale of this score is from 0 to 1.
% \begin{equation}
%     \ourtoolnospace\_Comparison\_Score(J^{1},P,J^{2})=\frac{1}{n}\sum_{i=1}^n (return\space1\spaceif is\_better(j_{i}^{1}),j_{i}^{2}),p_{i}) return\space0\space otherwise
%     \label{t2j:eq:comparison_score}
% \end{equation}

\textbf{Code generation as Baseline.} We perform the code translation process using the cheap LLM 4o-mini. This process leverages the standard prompt (see Appendix A.1) to translate input PyTorch code to JAX-generated code. We consider this configuration as baseline for \ourtool for comparison. 

\textbf{\ourtool's In-context learning for code generation.} We perform this process with following modules. First, the augmented prompt will be created with input information as the given Pytorch code snippet and the JSON format of other bug-solutions from other code samples of the problem-solving code dataset. Next, through the cheap LLM, the JAX generated code by \ourtool's approach is created. This process was done without the need of any fine-tuning steps, which are usually costly.

\textbf{Ground-truth JAX code generation.} We take advantage of costly LLM gpt-4o to generate the JAX code, called JAX initial code from given problem-solving code snippet. Next, the human bug fixing process was performed on this JAX initial code to make the JAX corrected code as ground truth. The main different between this process and the fixed bug dataset creation we mentioned earlier is that this module works with costly LLM. Finally, the output of these three modules will be use as the input of evaluation framework along with the PyTorch source code.

\subsection{Translation with different settings}
\textbf{Intrinsic Setting. }
In the intrinsic evaluation setting, we want to check if we can use knowledge from fixing 19 problem-solving code pairs of Pytorch and JAX correct code to improve the quality of cheap LLM to translate Pytorch code of the remaining sample. We perform this evaluation as a cross validation process over 20 samples of the fixed bug dataset. Next, we compare the output of three following modules.

\textbf{Extrinsic Setting. }
There are two main differences between intrinsic evaluation and extrinsic evaluation. First, for this set up we use the code snippets collected from another dataset, called Github dataset. We evaluate on 100 sample code snippets that are at repository level, meaning that the input code snippets, usually collected from single files, are not guarantee to be runnable and having test cases. Thus, we will use automated metrics we propose for the evaluation. Second, the extrinsic evaluation considered the output of costly LLMs, i.e. JAX translated code by this process) as the ground-truth data point.

\section{Experiment}
\subsection{Setup}
\textbf{Hardware Configuration.} For the human bug fixing process, software developers work in the Google Colab environment to debug and fix the code. They use Python 3 as a compiler and use one T4 GPU for running all sample code.

\textbf{Question-answering for cheap and costly LLMs.} For both cheap LLM (4o-mini) and costly LLM(gpt-4o), we perform the code generation process through the official interface of ChatGPT-pro. Each question will be created solely in a new topic, and from the answer given by ChatGPT's interface, we manually extract the code snippet as JAX-generated code. Other textual explanation in the output will be omitted. To add context to the existing prompt, we use the Upload function provided by ChatGPT's interface to assign a fixed bug dataset as a JSON file to our designed prompt.

\textbf{Executing Evaluation Prompts.} We leverage the access on OpenAI gpt-4o models by API key to get the scoring results for \ourtoolnospace\_CodeTrans\_Score. For \ourtoolnospace\_Comparison\_Score, we use ChatGPT-pro's interface to ask questions and receive answers. The reason for using ChatGPT-pro's interface instead of using API key is that some tasks of our work require JSON file upload.
% and for comparison score, we need to go manually through the output to see which code candidate is considered as better than the other models.

\subsection{Result}
\subsubsection{Translation Accuracy}
\begin{table}[t]
\small
\centering
\caption{Results with Cheap LLM (gpt-4o-mini) and Costly LLM (gpt-4o).}
\label{t2j:tbl:result}
\resizebox{\columnwidth}{!}{%
\begin{tabular}{lrrrr}
\toprule
\multirow{2}{*}{\textbf{Metrics}} & \multicolumn{2}{c}{\textbf{Intrinsic Evaluation}} & \multicolumn{2}{c}{\textbf{Extrinsic Evaluation}} \\
\cmidrule(lr){2-3} \cmidrule(lr){4-5}
 & \textbf{Baseline} & \textbf{\ourtoolnospace} & \textbf{Baseline} & \textbf{\ourtoolnospace} \\
\midrule
CodeBLEU                    & 0.19  & 0.29  & 0.41  & 0.38 \\
\ourtoolnospace\_CodeTrans\_Use.\_Ref         & 1.75  & 2.55  & 2.74  & 2.94 \\
\ourtoolnospace\_CodeTrans\_Func.\_Ref        & 0.35  & 1.3  & 2.43  & 3.02 \\
\ourtoolnospace\_CodeTrans\_Use.\_NoRef  & 1.60  & 2.45  & 2.81  & 2.98 \\
\ourtoolnospace\_CodeTrans\_Func.\_NoRef & 0.70  & 2.15  & 2.74  & 3.37 \\
\ourtoolnospace\_FixCost\_Score                & 163   & 87    & N/A   & N/A  \\
\ourtoolnospace\_Comparison\_Score           & 0  & 1  & 0.18  & 0.82 \\
\bottomrule
\end{tabular}
}
\end{table}

\begin{table}[t]
\small
\centering
\caption{Correlation of other metrics with \ourtoolnospace\_FixCost\_Score on Fixed Bug Dataset.}
\label{t2j:tbl:correlation}
\resizebox{\columnwidth}{!}{%
\begin{tabular}{lrr}
\toprule
\multirow{2}{*}{\textbf{Metric}} & \multicolumn{2}{c}{\textbf{Correlation with \ourtool\_FixCost\_Score}} \\
\cmidrule(lr){2-3}
 & \textbf{Pearson} & \textbf{Spearman} \\
\midrule
CodeBLEU                    & 0.04  & 0.2  \\
\ourtoolnospace\_CodeTrans\_Use.\_Ref         & 0.2  & 0.25  \\
\ourtoolnospace\_CodeTrans\_Func.\_Ref        & 0.07  & 0.07  \\
\ourtoolnospace\_CodeTrans\_Use.\_NoRef  & 0.09  & 0.19  \\
\ourtoolnospace\_CodeTrans\_Func.\_NoRef & 0.11  & 0.29 \\
\bottomrule
\end{tabular}}
\end{table}

% The results for PyTorch-to-JAX translation are shown in Figure~\ref{t2j:tbl:result}. 
% For intrinsic evaluation on 20 problem-solving code snippets, the baseline model achieves a CodeBLEU score of 0.19 (on a 0–1 scale). With \ourtoolnospace, 
From Figure~\ref{t2j:tbl:result}, our pipeline improves the CodeBLEU score to 0.29, representing a 10\% relative gain. In terms of \ourtoolnospace\_CodeTrans\_Use\_Ref, the augmented prompt yields an improvement of 0.8 points over the baseline. For functional correctness, \ourtool achieves an improvement of 0.95 point. Under the no-reference configuration—where the LLM evaluates only by comparing the input and translated code—\ourtool still delivers gains of 0.85 and 1.15 points for usefulness and functional correctness, respectively, as measured by the \ourtoolnospace\_CodeTrans metrics. Regarding the \ourtoolnospace\_Comparison\_Score, 100\% of the translations generated by \ourtool are judged superior to the baseline outputs. Finally, in terms of fixing cost, \ourtool enables GPT-4o-mini to produce code requiring only 87 fixing steps—roughly half the effort compared to fixing the baseline JAX outputs.
% The result for Pytorch-to-JAX translation is shown in Figure \ref{t2j:tbl:result}. We see that for intrinsic evaluation on 20 problem solving code snippets, the baseline model can achieve 0.19 (in the scale 0-1) in term of CodeBLEU score. With \ourtoolnospace, our pipeline is able to improve the CodeBLEU score as to 0.29, thus bring 10\% of improvement. In term of \ourtoolnospace\_CodeTrans\_Use\_Ref, our augmented prompt can bring 0.65 point as improvement over the baseline. In term of functional correctness, \ourtool achieves 1 point of improvement. With the no reference configuration, meaning the LLM will compare between the input and translated code only, \ourtool can still achieve 0.85 and 1.15 as improvement over the usefulness and functional correctness evaluated by \ourtoolnospace\_CodeTrans metrics. In term of \ourtoolnospace\_Comparison\_Score, over 75\% of translated code snippets by \ourtool are considered as better code than the baseline configuration. In terms of fixing cost, \ourtool can help 4o-mini generating the code set that only required 87 fixing steps, which is about half the effort for fixing the code set provided by baseline's JAX code generation process.

The extrinsic evaluation on the GitHub PyTorch code also highlights the superiority of \ourtool over the baseline in generating precise code. Interestingly, in this setting the baseline approach outperformed \ourtool by 3\% according to CodeBLEU. 
% A manual inspection of the generated cases indicates that \ourtool tends to produce longer and more detailed code.
% which can be penalized by textual and AST-based metrics such as CodeBLEU. 
In terms of the \ourtoolnospace\_CodeTrans metrics, our approach achieves improvements of up to 1.2 point in usefulness and 0.6 point in functional correctness. 
% Unlike the intrinsic evaluation, where reference and no-reference configurations achieved similar gains, the extrinsic evaluation shows greater improvements for \ourtool under the no-reference setting. This suggests that using initial translations from costly LLMs as ground truth may introduce noise into the evaluation process. Finally, the \ourtoolnospace\_Comparison\_Score shows significantly outperformance of \ourtool compared to baseline approach.
% The result from extrinsic evaluation on the Github Pytorch code also shows the superior of \ourtool compared to baseline in the ability of generating precise code. Interestlingly, for this setting the baseline approach outperformed \ourtool's result by 3\%. We manually check cases of code generation and see that the \ourtool's configuration tends to generate longer and more detailed code, which can be penalized by textual and AST matching scores like CodeBLEU. For \ourtoolnospace\_CodeTrans score, our approach achieve upto 1.1 point in usefulness and 1.3 point in functional correctness. Different from the intrinsic evaluation which the reference options achieve the same improvement with no reference option, \ourtool tends to have more improvement in terms of usefulness and functional correctness in No Reference configuration. It shows that there are possibilites that using initial translation result from costly LLMs as ground-truth might bring some noise to code evaluation accuracy. The \ourtool\_Comparison\_Score achieved the same improvement over baseline comapred to the intrinsic setting.
\begin{table}[]
\small
\centering
\caption{Comparison of running time (seconds) on Intrinsic Evaluation.}
\label{t2j:tbl:runningtime}
\begin{tabular}{lrrr}
\toprule
\textbf{PyTorch} & \textbf{Ground Truth} & \textbf{Baseline} & \textbf{\ourtoolnospace} \\
\midrule
1003  & 851 & 1232.9 & 449 \\
\bottomrule
\end{tabular}
\end{table}

\subsubsection{Correlation of Code Translation Metrics vs Human Fixing Cost}
% Since the human fixing cost is manually produced by software professionals and calculated without any involvement of LLMs, we consider this score as the pivot metric that reflects how far a buggy code snippet must be modified to obtain a correct translation. 
% We conduct an analysis base on the baseline's initial result and the cost to fix the generated code to correct code. 
From Table~\ref{t2j:tbl:correlation}, we observe that the \ourtool\_CodeTrans\_Score metrics show the strongest correlation with \ourtool\_FixCost\_Score under both Pearson and Spearman measures. 
% The Comparison Score could not be ranked since it is binary and, in the baseline setting, always returned zero 
% (i.e., GPT-4o judged all baseline outputs worse than those from \ourtoolnospace)
Overall, all metrics exhibit weak correlation (below 0.3) with fixing cost. 
One possible reason is that other metrics are continuous, whereas fixing cost is measured as discrete steps. 
\subsubsection{How close JAX-generated code by costly LLM is to being correct}
\begin{table}[t]
\small
\centering
\caption{Comparison of human fixing costs between baseline (cheap LLM with standard prompt) and JAX code initially generated by a costly LLM.}
\label{t2j:tbl:fixing_cost}
\resizebox{\columnwidth}{!}{%
\begin{tabular}{lrr}
\toprule
\textbf{Num. of Fixes} & \textbf{Correcting Cheap LLM's Code} & \textbf{Correcting Costly LLM's Code} \\
\midrule
Minimum & 1    & 0   \\
Maximum & 32   & 12  \\
Mean  & 8.15 & 2.77 \\
Median & 5    & 2   \\
Total   & 163  & 61  \\
\bottomrule
\end{tabular}}
\end{table}
We further analyze the quality of JAX generated code by costly LLM by meassuring the fixing cost to get the JAX correct code by costly LLM as ground-truth code. The result, shows in Table \ref{t2j:tbl:fixing_cost}, shows that while it requires much less effort for correcting costly LLM's output than baseline's output, it still requires in total 61 fixing steps to get the correct code set. 
% On average, there are over 2 fixing steps required per each file to correcting gpt-4o's generated code. 
% It shows that while generated code by costly LLM can be used as ground-truth dataset, future research should consider proposing approach for improve costly LLM quality to get the total correct code.
\subsubsection{Analysis on Running Time}
We analyze the running time of the corrected code from 3 settings for intrinsic evaluation in Table \ref{t2j:tbl:runningtime}. Results show that \ourtoolnospace can provide significant improvement as 2.5 times faster than running the baseline's output. Details of running time can be seen in Appendix B.

% \subsection{}

\section{Related Work}
\textbf{Challenges of Code Translation.} 
% Approaches of Code-to-code translation by transformers and LLMs have been proposed and applied in popular programming languages since the last decade, summarized in the study of \citep{sun2025surveyneuralcodeintelligence}. 
% Before the era of LLMs, supervised transformers have been proposed thanks to the efforts of collecting parallel corpus such as CodeSearchNet \citep{husain2020codesearchnetchallengeevaluatingstate}. 
% This dataset was collected from Github repositories, with information of documentation and implementation at method level. 
% Other challenges of collecting a pair dataset are the need for multiple languages and multiple domains of input problems, which have been solved in some recent datasets such as CodeTransOcean \citep{yan2023codetransoceancomprehensivemultilingualbenchmark}. 
One of the important challenges of code translation is to provide a metric for comparison between predicted and expected code. Research works show that just simply comparing source code by traditional textual similarity scores is not efficient \cite{tran2019migration}. Instead, code metrics that included information of syntactic/ semantic similarity between code snippets have been proposed \citep{zhou2023codebertscoreevaluatingcodegeneration}. 
% Taking advantage of large scale repository which does not have pair versions of code snippets between source and target languages is another challenge. 
Another challenge is that collecting parallel corpus for code translation is very expensive and require human effort for verification \citep{husain2020codesearchnetchallengeevaluatingstate}.
% The main direction to solve it is using unsupervised learning, with the idea of training the model for auto-completing or auto-correcting the target language code from an incomplete version of target code snippet that contained keywords/ syntaxes of source language code. 
To evaluate unsupervised code translation's output, automated test cases generation approaches have been proposed
% can be used to evaluate whether or not the translated code performed in the same way with the input code  
\citep{roziere2022leveragingautomatedunittests,peng2024humanevalxlmultilingualcodegeneration}. 
% Despite there are many works on optimizing a code translation process, 
Finally, studies show that there are many types of bugs extracted from LLMs' generated code \citep{dinh2023largelanguagemodelscode,zhang2024systematicliteraturereviewlarge}.

% Mix between supervised and unsupervised approach
\textbf{Machine Learning-based Approaches for Code Translation.}  
Supervised methods for code translation are typically trained on well-established datasets such as CodeXGLUE~\citep{lu2021codexgluemachinelearningbenchmark}. Among these methods, BERT-based models have proven particularly effective not only for code-to-code translation but also for a wide range of code generation tasks~\citep{guo2021graphcodebertpretrainingcoderepresentations,guo2022unixcoderunifiedcrossmodalpretraining,wang2023codet5opencodelarge,ahmad2021unifiedpretrainingprogramunderstanding}. 
% Their strength lies in a two-stage process: pre-training on large-scale code corpora to learn token-level mappings between source and target languages, followed by fine-tuning to adapt to specific translation tasks or datasets. 
Unsupervised code translation typically relies on transforming the source code into an intermediate representation (IR), followed by learning to generate target language code from that intermediate form. ~\cite{szafraniec2023codetranslationcompilerrepresentations} proposed \textit{Transcoder-IR}, a system that uses IR as a pivot language to translate between widely-used languages such as Java, Python, and C++. 
% Their work builds upon TransCoder~\citep{roziere2020unsupervised}, which learns translation mappings from incomplete code snippets in the target language without requiring parallel corpora. 
% ~\cite{roziere2022leveragingautomatedunittests} introduced an enhancement via automated unit testing, applying a back-translation strategy in which generated target code is translated back to the source language and evaluated on functional equivalence using the same test cases. 
% This approach enforces behavioral consistency as a criterion for translation quality. 
~\cite{huang2023programtranslationcodedistillation} proposed \textit{Codist}, which adopts a filtered IR to improve the precision of code translation through a process called code distillation. 
 ~\cite{tehranijamsaz2024coderosetta} presented \textit{CodeRosetta}, a framework for unsupervised translation from C to CUDA. Their method exploits the syntactic similarity between the two languages, leveraging abstract syntax trees (ASTs) as the pivot representation to learn structural correspondences. ~\cite{roziere2021dobfdeobfuscationpretrainingobjective} emphasized the significance of a pre-training objective based on recovering broken or obfuscated code. 

\section{Limitations}
% We identify several limitations of our work. 
First, \ourtool has not yet been applied to improving open LLMs, due to budget constraints that limit our ability to hire software professionals for the human bug-fixing process on these models. Second, our measure of fixing cost is currently based only on counting fixing steps, whereas in practice, each fix may vary in difficulty. 
% To improve this, 
There is a need of an algorithm that estimates the relative effort of each bug-fixing step. 
Third, we conducted the human bug-fixing process only on the problem-solving code dataset, which we want to extend this process for other domains. 

% In future versions of \ourtool, we plan to collect bug datasets from other general domains.

% We identify several limitations of our work. First, the current version of T2J hasn't been applied for improving open LLMs, given the restriction of our budget for hiring software professional to perform the human bug fixing process on open LLMs. Second,we currently leverage the fixing cost by the number of fixes, while in fact each fix might have different difficult level. One future work that can alleviate this drawback is to design an algorithm to judge the effort of bug fixing for each step. Third, we only perform the human bug fixing process on the problem-solving dataset. In future version of \ourtool, we will extend the scope of human bug fixing to cover datasets in other general domaina.
\section{Conclusion and Future Works}
In this work, we show that \ourtool\ can achieve significant improvement compared to baselines as original 4o-mini model for PyTorch-to-JAX code translation. In future work, we attempt to apply our approach for newer open-source LLMs and leverage more advance techniques as supervised fine-tuning and direct preference optimization. 
% to guide LLMs to improve their abilities of auto-fixing bugs from generated code, inspired by the way of human-in-the-loop code verification that we proposed in this work.

\newpage
\bibliography{custom}
\bibliographystyle{icml2026}

%%%%%%%%%%%%%%%%%%%%%%%%%%%%%%%%%%%%%%%%%%%%%%%%%%%%%%%%%%%%%%%%%%%%%%%%%%%%%%%
%%%%%%%%%%%%%%%%%%%%%%%%%%%%%%%%%%%%%%%%%%%%%%%%%%%%%%%%%%%%%%%%%%%%%%%%%%%%%%%
% APPENDIX
%%%%%%%%%%%%%%%%%%%%%%%%%%%%%%%%%%%%%%%%%%%%%%%%%%%%%%%%%%%%%%%%%%%%%%%%%%%%%%%
%%%%%%%%%%%%%%%%%%%%%%%%%%%%%%%%%%%%%%%%%%%%%%%%%%%%%%%%%%%%%%%%%%%%%%%%%%%%%%%
\newpage
\appendix
\onecolumn
\appendix
\section*{Appendix}
\addcontentsline{toc}{section}{Appendix}

\section{Prompt Template}
\label{t2j:appx:prompt_template}
\subsection{Standard Prompt for PyTorch-to-JAX translation}
\begin{figure}[ht]
\centering
\begin{tcolorbox}[colback=black!5!white, colframe=black!75!black]
You are an expert in programming language translation from PyTorch to JAX. In this task, I
will give you input as PyTorch code. Please translate this input PyTorch code to JAX code:

{\color{blue}Input Source Code Snippet:}
            {\color{brown} \{CODE\}}
    
\end{tcolorbox}
\caption{ Standard Prompt for PyTorch-to-JAX code translation. {\color{blue} The prompt in blue shows the immediate query after the prompt}. {\color{brown} \{CODE\}is the starting string of the code.}}
\label{t2j:prompt:standard}
\end{figure}

\subsection{Augmented Prompt for PyTorch-to-JAX translation}
\begin{figure}[ht]
\centering
\begin{tcolorbox}[colback=black!5!white, colframe=black!75!black]
You are an expert in programming language translation from PyTorch to JAX. In this task, I
will give you two inputs:

1. Pytorch source code.

2. A JSON file that contains a dataset of common errors in PyTorch-to-JAX translation by Weak
LLM 4o-mini. Each data point contains the following fields:

- \texttt{Example\_id}: ID of the source code.

- \texttt{Input\_Code}: Source code in Pytorch.

- \texttt{LLM\_weak\_output}: JAX translated code of \texttt{Input\_Code} using a weak LLM ($4o$-mini).
\texttt{LLM\_fix\_output}: Fixed JAX code from \texttt{LLM\_weak\_output} by the process of manually check and fix errors conducted by software developers.

- \texttt{Errors}: This is a list of errors that appeared in the process of manually checking and
fixing bugs from \texttt{LLM\_weak\input}. Each error item has the following labels:

• \texttt{"Error\_Code"}: The part of \texttt{LLM\_weak\_output} that caused the error.

• \texttt{"Error"}: the error message returned by compilation.

• \texttt{"Fix\_info"}: the textual description of how to fix the error code

• \texttt{"Fixed\_Code"}: The fixed code corresponding to the \texttt{"Error\_Code"} part.

3. The data.csv file thich stored possible input when running some examples in the JSON file.
Your task is to reason and get the output JAX code from these above inputs. Please note that
you can learn the process of error fixing in Torch-to-JAX translation in 2) JSON file. Now I
will give you a set of input in the next query.\\

{\color{blue}Input Source Code Snippet:}
            {\color{brown} \{CODE\}}
    
\end{tcolorbox}
\caption{ Prompt for Augmenting to the weak LLM. {\color{blue} The prompt in blue shows the immediate query after the prompt}. {\color{brown} \{CODE\}is the starting string of the code.}}
\label{t2j:prompt:automated}
\end{figure}

\subsection{Evaluation Prompt for \ourtoolnospace\_{CodeTrans}\_{Score}}
\subsubsection{Functional Correctness}
See the prompt without reference at Figure \ref{t2j:prompt:codetrans:func:noref} and the prompt with reference at Figure \ref{t2j:prompt:codetrans:func:ref}.
\begin{figure}[ht]
\centering
\begin{tcolorbox}[colback=black!5!white, colframe=black!75!black]
You will be given a JAX code snippet that was translated from PyTorch source code.  
Your task is to rate the snippet on **one metric only**: its **functional correctness**.

Please ensure you read and understand these instructions carefully before reviewing.
Refer to this guide as needed during the evaluation process.

Evaluation Criteria:  

Functional Correctness (0–4) — How well the JAX code preserves the behavior of the original PyTorch code.  

You are to assess whether the JAX code would produce equivalent outputs to the original PyTorch code across possible inputs, even though the PyTorch code is not shown. Consider unit-test-style logic and general expectations of equivalence.

- A score of 0: The translation is completely incorrect and meaningless.

- A score of 4: The translation is fully correct and handles all core functionalities as expected.

Evaluation Steps:

1. Assume the code was translated from PyTorch and should preserve its logic.

2. Evaluate whether the JAX code appears complete, meaningful, and implementationally correct based on general expectations for such translations.

3. Assign a score for functional correctness on a scale from 0 to 4.

Input Source Code in PyTorch:

{\color{brown} \{SOURCE\_CODE\}}

Translated JAX Code Snippet:

{\color{brown} \{TRANSLATED\_CODE\}}

Evaluation Form:  

Functional Correctness (scores ONLY):
\end{tcolorbox}
\caption{ Prompt for Scoring Functional Correctness by \ourtool\_CodeTrans\_Func\_NoRef}
\label{t2j:prompt:codetrans:func:noref}
\end{figure}

\begin{figure}[ht]
\centering
\begin{tcolorbox}[colback=black!5!white, colframe=black!75!black]
You will be given a JAX code snippet that was translated from PyTorch source code.  
Your task is to rate the snippet on **one metric only**: its **functional correctness**.

Please ensure you read and understand these instructions carefully before reviewing.
Refer to this guide as needed during the evaluation process.

Evaluation Criteria:  
Functional Correctness (0–4) — How well the JAX code preserves the behavior of the original PyTorch code.  

You are to assess whether the JAX code would produce equivalent outputs to the original PyTorch code across possible inputs, even though the PyTorch code is not shown. Consider unit-test-style logic and general expectations of equivalence.

- A score of 0: The translation is completely incorrect and meaningless.

- A score of 4: The translation is fully correct and handles all core functionalities as expected.

Evaluation Steps:

1. Assume the code was translated from PyTorch and should preserve its logic.

2. Evaluate whether the JAX code appears complete, meaningful, and implementationally correct based on general expectations for such translations.

3. Assign a score for functional correctness on a scale from 0 to 4.

Input Source Code in PyTorch:

{\color{brown} \{SOURCE\_CODE\}}

Translated JAX Code Snippet:

{\color{brown} \{TRANSLATED\_CODE\}}

Reference JAX Code Snippet:

{\color{brown} \{REFERENCE\}}

Evaluation Form:  

Functional Correctness (scores ONLY):
\end{tcolorbox}
\caption{ Prompt for Scoring Functional Correctness by \ourtool\_CodeTrans\_Func\_Ref}
\label{t2j:prompt:codetrans:func:ref}
\end{figure}

\subsubsection{Usefulness}
See the prompt without reference at Figure \ref{t2j:prompt:codetrans:usefulness:noref} and the prompt with reference at Figure \ref{t2j:prompt:codetrans:usefulness:ref}.
\begin{figure}[ht]
\centering
\begin{tcolorbox}[colback=black!5!white, colframe=black!75!black]
Your task is to rate the snippet on **one metric only**: its **usefulness** for understanding and reusing the logic of a typical PyTorch implementation.

Please ensure you read and understand these instructions carefully before reviewing.
Refer to this guide as needed during the evaluation process.

Evaluation Criteria:  
Usefulness (0–4) — How useful the JAX code is for replicating or adapting the functionality of a typical PyTorch source code implementation.

- A score of 0: The JAX translated snippet is irrelevant or confusing and does not help at all.

- A score of 1: The JAX translated snippet includes some related elements but is mostly unhelpful.

- A score of 2: The JAX translated snippet is somewhat useful but needs substantial modification.

- A score of 3: The JAX translated snippet is helpful with minor revisions needed.

- A score of 4: The JAX translated snippet is very helpful and covers the intended functionality clearly.

Evaluation Steps:

1. Assume the PyTorch source code performs a well-defined functionality.

2. Determine whether the JAX translated code snippet enables meaningful reuse or guidance toward equivalent implementation.

3. Assign a score for usefulness from 0 to 4.

Input Source Code in PyTorch:  

{\color{brown} \{SOURCE\_CODE\}}

Translated JAX Code Snippet:

{\color{brown} \{TRANSLATED\_CODE\}}

Evaluation Form:  

Usefulness (scores ONLY):
\end{tcolorbox}
\caption{ Prompt for Scoring Usefulness by \ourtool\_CodeTrans\_Use\_NoRef}
\label{t2j:prompt:codetrans:usefulness:noref}
\end{figure}

\begin{figure}[ht]
\centering
\begin{tcolorbox}[colback=black!5!white, colframe=black!75!black]
Your task is to rate the snippet on **one metric only**: its **usefulness** for understanding and reusing the logic of a typical PyTorch implementation.

Please ensure you read and understand these instructions carefully before reviewing.
Refer to this guide as needed during the evaluation process.

Evaluation Criteria:  
Usefulness (0–4) — How useful the JAX code is for replicating or adapting the functionality of a typical PyTorch source code implementation.

- A score of 0: The JAX translated snippet is irrelevant or confusing and does not help at all.

- A score of 1: The JAX translated snippet includes some related elements but is mostly unhelpful.

- A score of 2: The JAX translated snippet is somewhat useful but needs substantial modification.

- A score of 3: The JAX translated snippet is helpful with minor revisions needed.

- A score of 4: The JAX translated snippet is very helpful and covers the intended functionality clearly.

Evaluation Steps:

1. Assume the PyTorch source code performs a well-defined functionality.

2. Determine whether the JAX translated code snippet enables meaningful reuse or guidance toward equivalent implementation.

3. Assign a score for usefulness from 0 to 4.

Input Source Code in PyTorch:  

{\color{brown} \{SOURCE\_CODE\}}

Translated JAX Code Snippet:

{\color{brown} \{TRANSLATED\_CODE\}}

Reference JAX Code Snippet:

{\color{brown} \{REFERENCE\}}

Evaluation Form:  

Usefulness (scores ONLY):
\end{tcolorbox}
\caption{ Prompt for Scoring Usefulness by \ourtool\_CodeTrans\_Use\_Ref}
\label{t2j:prompt:codetrans:usefulness:ref}
\end{figure}

\subsection{Evaluation Prompt for \ourtoolnospace\_{Comparison}\_{Score}}
The prompt for querying the \ourtoolnospace\_{Comparison}\_{Score} can be seen in Figure \ref{t2j:prompt:comparison}.
\begin{figure}[ht]
\centering
\begin{tcolorbox}[colback=black!5!white, colframe=black!75!black]
You are an expert in PyTorch to JAX translation. I provide 3 inputs: 
1 . PyTorch input code; 2. Translated Code Candidate A; 3. Translated Code Candidate B. Which candidate is a better translation result for this Pytorch code. 

Input Pytorch code:

{\color{brown} \{CODE\}}

2. Translated Code A:

{\color{brown} \{TRANSLATE\_CODE\_A\}}

3. Translated Code B:

{\color{brown} \{TRANSLATE\_CODE\_B\}}

Please also provide the reason why you consider a candidate better than the other translated code candidate.
    
\end{tcolorbox}
\caption{ Prompt for \ourtool\_Comparison\_Score.}
\label{t2j:prompt:comparison}
\end{figure}

\section{Additional Results}
\subsection{Analysis on Categories of Bugs}
We perform a study on the categorization of bugs on the fixed bug dataset as following. First, two software professionals will go through all the bugs and discuss about the categorizations. Second, from this categorization, they go to the dataset's entities for the second time and do the annotation for categories. We summarize the categorization in Table \ref{t2j:tbl:category_bug}. We further classify types of bugs for some categories to sub-categories, shown in Table \ref{t2j:tbl:sub_categorization} and Table \ref{t2j:tbl:other_error}. We upload each case of this categorization process in the replication package.

\begin{table}[t]
\centering
\caption{Error categories and their counts in Human Bug Fixing dataset.}
\label{t2j:tbl:category_bug}
\begin{tabular}{lp{2cm}}
\toprule
\textbf{Error Main Category} & \textbf{Count} \\
\midrule
{\color[HTML]{FF0000}\textbf{Training loops, training steps, model fitting}}             & 32 \\
{\color[HTML]{0000FF}\textbf{Other miscellaneous}}                                       & 43 \\
{\color[HTML]{38761D}\textbf{Model definitions, LinearModel classes, encoders/decoders}} & 51 \\
\textbf{Loss functions, gradient computation, criterion}                                 & 12 \\
\textbf{JAX-specific constructs: jit, grad, PRNG, etc.}                                  & 18 \\
\textbf{Iteration patterns: for, while, data loops}                                      & 1 \\
\textbf{Parameter updates}                                                               & 3 \\
\textbf{Final layers, return statements, outputs}                                        & 3 \\
\midrule
\textbf{TOTAL}                                                                           & 163 \\
\bottomrule
\end{tabular}
\end{table}

\begin{table}[t]
\centering
\caption{Error subcategories under \textbf{training loops, training steps, and model fitting}.}
\label{t2j:tbl:sub_categorization}
\begin{tabular}{lr}
\toprule
\textbf{Error Subcategory} & \textbf{Count} \\
\midrule
Misc training issues                                                   & 8 \\
Improper passing/using \texttt{rng\_key}/\texttt{prng\_key}            & 6 \\
Epoch in \texttt{range(...)} loop issues                               & 3 \\
Incorrect usage of Flax \texttt{TrainState} and \texttt{state.apply\_gradients} & 3 \\
Incorrect usage of wrappers (e.g., \texttt{train\_model(...)} / \texttt{fit(...)}) & 3 \\
Train steps return only new state/params without loss at epoch level   & 3 \\
JIT/static argument handling for training functions                    & 3 \\
Errors with batches or dataloaders in training                         & 1 \\
Loop constructs that break vectorization                               & 1 \\
Optimizer update/apply patterns in the training loop                   & 1 \\
\bottomrule
\end{tabular}
\end{table}

\begin{table}[t]
\centering
\caption{Error subcategories under \textbf{other miscellaneous}.}
\label{t2j:tbl:other_error}
\begin{tabular}{lr}
\toprule
\textbf{Error Subcategory} & \textbf{Count} \\
\midrule
Data arrays, tensors, and dataset values (e.g., creating arrays, specifying shapes) & 8 \\
Dot products with parameters (e.g., \texttt{params["w"]})                            & 5 \\
Initialization, often in class constructors (\texttt{\_\_init\_\_})                 & 1 \\
Dot products, sums, or nonlinear transforms                                         & 3 \\
Neural network layers and activations (e.g., \texttt{nn.relu}, \texttt{nn.Dense}, LSTMs, decoders/encoders) & 5 \\
Tensor dimension errors                                                             & 2 \\
Generating synthetic data for CSV                                                   & 1 \\
Errors with constant declaration (e.g., epoch)                                      & 1 \\
Incomplete functions/placeholders                                                   & 17 \\
\bottomrule
\end{tabular}
\end{table}

\subsection{Running time Analysis}
We perform the running process on a T4 GPU for all the code. We set the timeout of program to run as 180 seconds. Results for each sample in the intrinsic evaluation are shown in Table \ref{t2j:tbl:runtime_details}.

\begin{table}[t]
\centering
\caption{Example results comparing PyTorch, ground truth, baseline, and \ourtoolnospace\ (T2J) outputs.}
\label{t2j:tbl:runtime_details}
\begin{tabular}{lrrrr}
\toprule
\textbf{Example ID} & \textbf{PyTorch} & \textbf{Ground Truth} & \textbf{Baseline} & \textbf{T2J} \\
\midrule
e1   & 6.61   & 12.7   & 9.78   & 8.79   \\
e2   & 10.00  & 60.0   & 24.4   & 3.59   \\
e3   & 8.57   & 17.3   & 4.14   & 5.18   \\
e4   & 8.82   & 22.1   & 21.8   & 5.36   \\
e5   & 8.48   & 35.0   & 35.2   & 6.53   \\
e6   & 13.0   & 24.0   & 3.0    & 13.68  \\
e7   & 8.23   & 5.0    & 90.0   & 14.53  \\
m1   & 17.0   & 26.0   & 180.0  & 15.62  \\
m3   & 180.0  & 180.0  & 4.0    & 1.05   \\
m4   & 180.0  & 18.0   & 180.0  & 29.0   \\
m5   & 102.0  & 57.0   & 180.0  & 19.5   \\
m6   & 18.0   & 6.0    & 61.0   & 4.0    \\
m7   & 78.0   & 65.0   & 47.9   & 62.0   \\
m8   & 240.0  & 19.0   & 11.8   & 10.0   \\
h1   & 0.48   & 4.0    & 2.5    & 10.0   \\
h3   & 97.0   & 7.0    & 44.1   & 19.0   \\
h4   & 31.1   & 180.0  & 180.0  & 15.0   \\
h5   & 19.6   & 180.0  & 180.0  & 42.0   \\
h6   & 0.31   & 9.0    & 4.0    & 180.0 \\
h10  & 0.89   & 14.0   & 7.6    & 2.0    \\
\midrule
Total & 1028.09 & 941.1 & 1271.22 & 466.83 \\
\bottomrule
\end{tabular}
\end{table}
\section{Configurations}
For the task that required user interface action with LLMs, we use the default ChatGPT-pro setting for gpt-4o and 4o-mini. Most of the data were created before July 31st, 2025 when 4o-mini was still available on ChatGPT's interface. For task like LLM-based metric calculation, we leverage OpenRouter's API\footnote{https://openrouter.ai/} to perform the implementation of these tasks. We also report the hyper parameters for querying costly LLMs for code evaluation in the replication packages.

\end{document}